\documentclass[runningheads]{llncs}
\usepackage[table]{xcolor}
\usepackage{graphicx}
\usepackage{wrapfig}
\usepackage{tabularx}
\usepackage{arydshln}
\usepackage{subcaption}
\usepackage[export]{adjustbox}
\usepackage{multirow}
\usepackage{dsfont}
\usepackage{xcolor,colortbl}
\usepackage{amsmath,amsfonts,bm,mathtools, bbm}
\usepackage{amssymb}
\usepackage{cite}
\usepackage{hyperref}
\usepackage{amssymb}
\usepackage{enumitem}
\usepackage{multirow}
\usepackage{tabularray}
\usepackage{booktabs}
\usepackage{tabularx}
\usepackage{arydshln}
\usepackage{orcidlink}
\usepackage{pifont}

\usepackage{xcolor, amsmath}
\definecolor{amaranth}{rgb}{0.9, 0.17, 0.31}
\colorlet{green}{green!20}
\colorlet{yellow}{yellow!60}
\colorlet{red}{red!30}
\definecolor{LightBlue}{RGB}{220, 220, 235} 
\definecolor{DarkGreen}{RGB}{1, 100, 32}

\usepackage{soul}
\usepackage{setspace}
\usepackage[linesnumbered,ruled,vlined]{algorithm2e}

\SetKwComment{Comment}{\color{green!80!black!190}// }{}

\SetKwProg{Function}{function}{}{}
\SetKw{Continue}{continue}

\DontPrintSemicolon
\SetAlFnt{\small}
\SetAlgorithmName{Pseudocode}{algorithmautorefname}

\usepackage{tikz}
\usetikzlibrary{fit,calc}

\newcommand{\boxitt}[2]{
    \tikz[remember picture,overlay] \node (A) {};\ignorespaces
    \tikz[remember picture,overlay]{\node[yshift=3pt,fill=#1,opacity=.25,fit={($(A)+(0,0.15\baselineskip)$)($(A)+(.858\linewidth,-{#2}\baselineskip - 0.25\baselineskip)$)}] {};}\ignorespaces
}

\usepackage{algpseudocode}

\begin{document}
%

\title{Boosting Vision-Language Models for Histopathology Classification: Predict all at once} 

\titlerunning{Boosting VLMs for Histopathology Classification}
%
%
\author{Maxime Zanella\inst{1,2}\thanks{M. Zanella and F. Shakeri—Equal contribution.} \orcidlink{0009-0009-4030-3704} 
\and
Fereshteh Shakeri\inst{3,4}$^*$ \orcidlink{0009-0006-4830-9631} 
\and
Yunshi Huang \inst{3,4} \orcidlink{0000-0003-1636-7248} 
\and \\
Houda Bahig \inst{4} \orcidlink{0000-0002-5607-6368} 
\and
Ismail Ben Ayed \inst{3,4} \orcidlink{0000-0002-9668-8027} 
}
\authorrunning{M. Zanella \textit{et al.}}
%
\institute{Université Catholique de Louvain (UCLouvain), Louvain-La-Neuve, Belgium \texttt{maxime.zanella@uclouvain.be} \and Université de Mons (UMons), Mons, Belgium \and \'Ecole de Technologie Supérieure (\'ETS), Montréal, Canada \texttt{fereshteh.shakeri.1@etsmtl.net} \and Centre de Recherche du Centre Hospitalier de l’Université de Montréal (CRCHUM), Montréal, Canada}
\maketitle
\begin{abstract}
The development of vision-language models (VLMs) for histo-pathology has shown promising new usages and zero-shot performances. However, current approaches, which decompose large slides into smaller patches, focus solely on inductive classification, \textit{i.e.,} prediction for each patch is made independently of the other patches in the target test data. We extend the capability of these large models by introducing a transductive approach. By using text-based predictions and affinity relationships among patches, our approach leverages the strong zero-shot capabilities of these new VLMs without any additional labels. Our experiments cover four histopathology datasets and five different VLMs. Operating solely in the embedding space (\textit{i.e.,} in a black-box setting), our approach is highly efficient, processing $10^5$ patches in just a few seconds, and shows significant accuracy improvements over inductive zero-shot classification. Code available at \url{https://github.com/FereshteShakeri/Histo-TransCLIP}.

\end{abstract}

\section{Introduction}
\label{sec:intro}
 Histology slides obtained from Whole Slide Image (WSI) \textbf{\cite{pantanowitz2010digital}} scanners play a crucial role in cancer diagnosis and staging \cite{madabhushi2009digital}. These slides offer a detailed view of diseased tissues, aiding in the determination of treatment options. Pathologists primarily diagnose cancers by examining WSIs to identify different tissue types. However, manually analyzing these WSIs imposes a significant workload, leading to substantial delays in reporting time. Moreover, in real clinical environments, the classification of cancer-related tissues is highly diverse, encompassing various cancer sites. Even within a single cancer site, tasks can vary in their levels of class granularity. Therefore, automating tissue-type classification in histology images (\cite{komura2018machine, petushi2006large, qureshi2008adaptive, tabesh2007multifeature, bilgin2007cell} to list a few) holds significant clinical value but is hindered by the difficulty of collecting large labeled datasets and the variability of fine-grained labels. \\

\par

The advent of multi-modal learning methods that process and integrate information from diverse modalities has alleviated some issues of training fine-grained classifiers and collecting costly labeled data. In particular, vision-language models (VLMs) such as CLIP \cite{radford2021clip} and ALIGN \cite{jia2021scaling} have gained popularity in computer vision, and demonstrated promising generalization capabilities across various downstream tasks. These so-called foundation models jointly train vision and text embeddings using contrastive learning on large-scale image-text datasets. This new multi-modal paradigm can naturally be extended to clinical scenarios, where combinations of multiple data modalities--mainly texts and images--are often adopted to obtain more accurate and comprehensive diagnosis. For example, clinical notes and pathology reports, alongside histopathology slides, are commonly used for throughout analysis \cite{Iryna2024}. However, the direct application of deep learning techniques, more specifically vision-language pre-training strategies, to medical imaging is complex, due to the lack of fine-grained expert medical knowledge, which is required to capture specialized information \cite{Chen2022}. This issue has been partly addressed for histopathology slides by collecting diverse data from scientific publications, Twitter, or even YouTube videos \cite{plip, ikezogwo2023quilt, lu2024avisionlanguage}.\\
\par

\begin{figure}
    \centering
    \begin{subfigure}{\textwidth}
        \includegraphics[width=\textwidth]{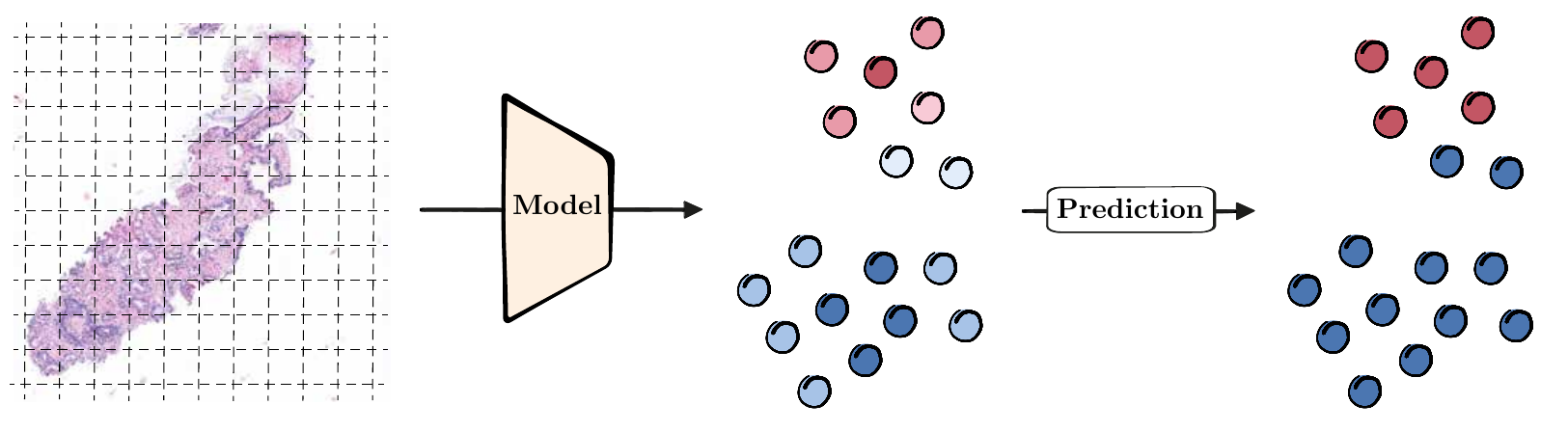}
        \caption{In the typical inductive setting, a model is trained and then used to infer on each patch separately. This approach can be efficient when large annotated datasets for each task are available. This procedure often involves predicting the most probable class (argmax).}
    \end{subfigure}

    \begin{subfigure}{\textwidth}
        \includegraphics[width=\textwidth]{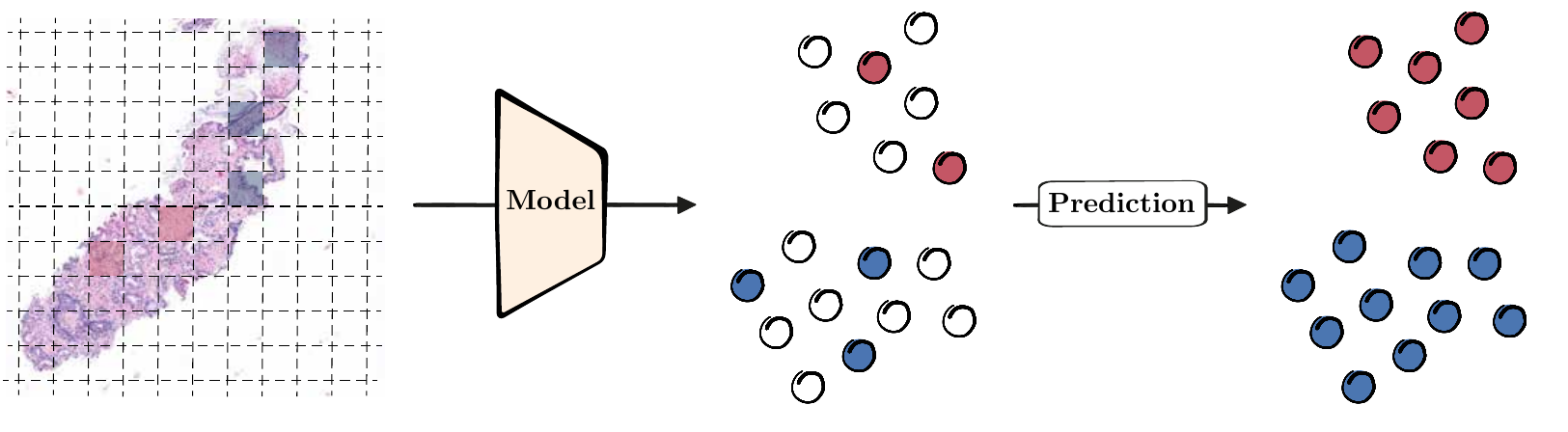}
        \caption{In the traditional transductive few-shot setting, a pre-trained encoder (e.g., on ImageNet or large-scale histology dataset) requires manual annotations for the new task to propagate information from labeled to unlabeled samples. This process often involves measuring affinities or distances between encoded samples.}
    \end{subfigure}

    \begin{subfigure}{\textwidth}
        \includegraphics[width=\textwidth]{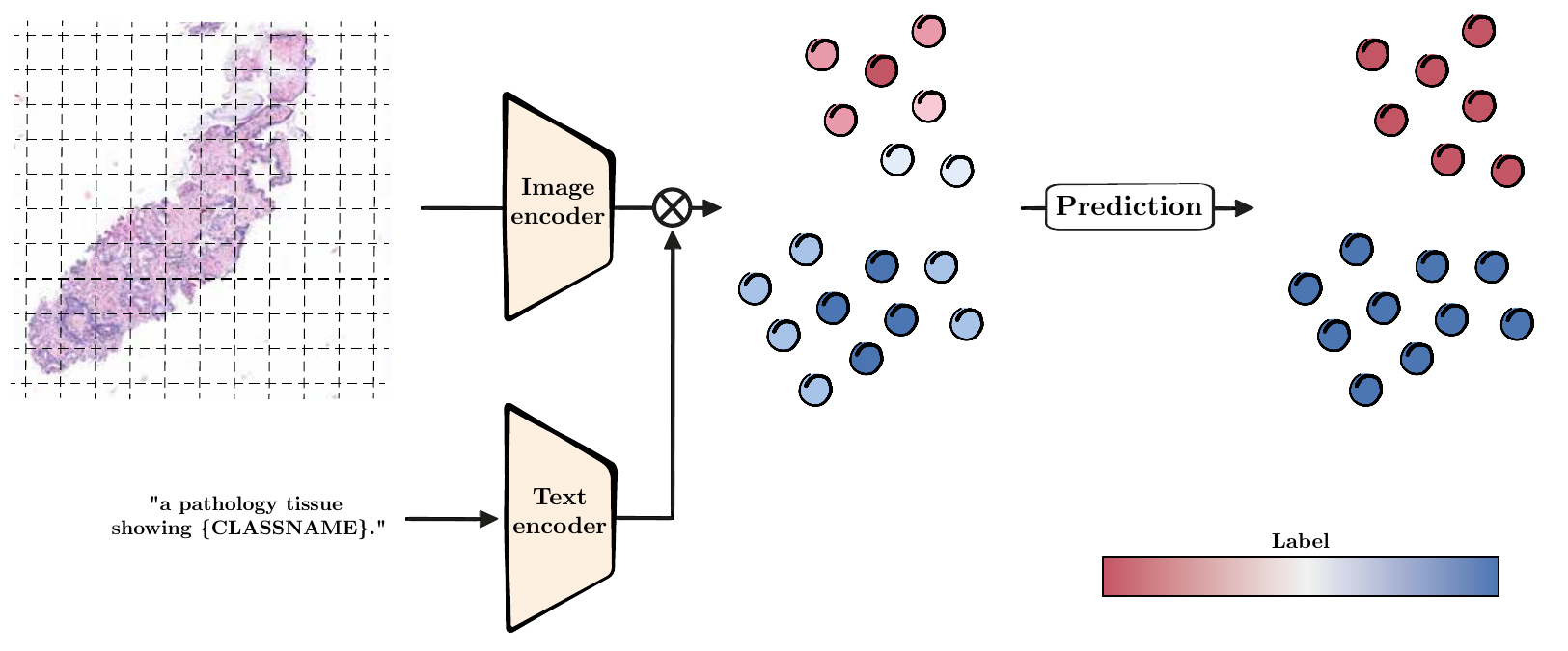}
        \caption{VLMs leverage textual descriptions of each class to generate pseudo-labels without any manual annotation. These initial predictions can then be refined, for example, by leveraging the data structure.}
        \label{subfig:transclip}
    \end{subfigure}
    \caption{Illustration depicting histopathology classification in the inductive setting (a), the commonly-used few-shot transductive setting (b), and the zero-shot transductive setting enabled by VLMs (c).}
    \label{fig:transduction}
\end{figure}

Current usage of such models predominantly align with the {\em inductive} paradigm, i.e., inference for each test sample is performed independently from the other samples within the test dataset. In contrast, {\em transduction} performs joint inference on all the test samples of a task, leveraging the statistics of the target 
unlabeled data~\cite{788640, joachims1999transductive}.
Transduction has primarily been explored for few-shot classification of natural images, to tackle the inherent challenges of training under limited supervision~\cite{dhillon2019baseline, boudiaf2020information}. These techniques utilize labeled samples to transfer information to unlabeled test data. Interestingly, in the novel multi-modal paradigm introduced by VLMs, supervision can be instead provided through textual descriptions of each classes (prompts), in a zero-shot setting, \textit{e.g.,} \texttt{a pathology tissue showing} [$\texttt{class name}$]. Along with their corresponding representation derived from the language encoder, similarities between text and image embeddings can be leveraged to enable transductive inference even in the {\em zero-shot} scenario, as pointed by recent works in computer vision \cite{zanella2024boosting, martin2024transductive} (see Figure~\ref{fig:transduction}).

\paragraph{\textbf{Contributions.}} With the ongoing development of foundation models in medical imaging and specifically histopathology, and the potential application of transductive inference, our objective is to improve zero-shot predictions of VLMs within this framework. Our main contributions can be summarized as follows: 
\begin{itemize}
\item We compare the zero-shot performance of vision-language models for histology and propose an effective transductive method to significantly boost their accuracy by leveraging the structure among patches during inference.
\item Our transductive approach does not require labels; instead, it utilizes text-based predictions as regularization.
\item To alleviate the computational workload, our method relies on the pre-computed features only, without access to the pre-trained weights, thus accommodating black-box constraints. This makes it feasible to process very large-scale slides in a matter of seconds.
\end{itemize}

\section{Related Work}
\label{sec:rw}
\paragraph{\textbf{VLMs for histology.}} Unlike natural images, which are often available in millions (\textit{e.g.,} CLIP \cite{radford2021clip} is trained on 400M image-text pairs), clinical image-text pairs are more challenging to amass. Similar to other works introducing VLMs for medical imaging (\textit{e.g.,} for radiology \cite{convirt,MedCLIP,medklip}, or ophthalmology \cite{FLAIR2023}), several VLMs for computational pathology has appeared recently, differentiating themselves primarily through their data collection and curation methodologies. PLIP \cite{plip} curates OpenPath, a large dataset of pathology images paired with text descriptions. Quilt-1M \cite{ikezogwo2023quilt} stands as one of the largest vision-language histopathology dataset to date, comprising 1 million image/text pairs sourced from YouTube videos. More recently CONCH \cite{lu2024avisionlanguage} integrates parts of the PubMed Central Open Access Dataset yielding 1.17 million samples. As these new VLMs have been developed in a short amount of time, determining the most suitable one is not straightforward. Therefore, we provide a comparison of these models and demonstrate the applicability of our approach across each of them.
\paragraph{\textbf{Transductive learning.}} In the few-shot literature solely based on vision models, transduction leverages both the few labeled samples and unlabeled test data~\cite{788640, joachims1999transductive}, outperforming inductive methods~\cite{dhillon2019baseline, boudiaf2020information, liu2020prototype, ziko2020laplacian}. This setting, widely explored in computer vision, has been recently deployed in histopathology   \cite{sadraoui2023transductive}, using the annotations of a few patches from slides of liver. 
However, previously mentioned transductive methods have been shown to suffer from significant performance drops when applied to VLMs \cite{martin2024transductive, zanella2024boosting}. This motivated a few, very recent transductive methods in computer vision, focusing on natural images and explicitly leveraging the textual modality along the image embeddings \cite{martin2024transductive, zanella2024boosting}. In contrast to \cite{sadraoui2023transductive}, our work exploits the findings and transductive-inference zero-shot objective in \cite{zanella2024boosting}, aiming to boost the predictive accuracy of pretrained histopathology VLMs without any supervision.

\section{Method}
\label{method}

In this section, we describe the Histo-TransCLIP objective function
for transductive inference in VLMs, for the $K$-class prediction problem. This objective function depends on two types of variables: (i) assignment variables ${\mathbf z}_i = (z_{i,k})_{1 \leq k \leq K} \in \Delta_K$, for each patch $i \in \cal Q$; and (ii) Gaussian Mixture Model (GMM) parameters $\boldsymbol{\mu} = (\boldsymbol{\mu}_k)_{1 \leq k \leq K}$ and $\boldsymbol{\Sigma}$. We will first detail the main components of Histo-TransCLIP, before deriving the overall procedure. 
\paragraph{\textbf{Gaussian modelization}} We modelize the likelihood of the target data as a balanced mixture of multivariate Gaussian distributions, each representing a class $k$, parameterized by mean vector $\boldsymbol{\mu}_k$ and a diagonal (shared among classes) covariance matrix $\boldsymbol{\Sigma}$:
\[p_{i,k}  \propto \det(\boldsymbol{\Sigma})^{-\frac{1}{2}} \exp \left(-{\frac {1}{2}}({\mathbf f}_i - \boldsymbol{\mu}_k)^\top \boldsymbol{\Sigma}^{-1}({\mathbf f}_i - \boldsymbol{\mu}_k) \right)\]
where ${\mathbf f}_i$ represents the encoding of patch $i$.
\paragraph{\textbf{Text-based predictions}.} When dealing with a zero-shot classification problem based on a VLM, and given a set of $K$ candidate classes, one can get textual embeddings ${\mathbf t}_k$ (\textit{e.g.,} from \texttt{a pathology tissue showing} [$\texttt{kth class name}$], $k=1,\dots, K$). Then pseudo-labels can be obtained by evaluating the softmax function of the cosine similarities between these two encoded modalities with $\tau$ being a temperature parameter:
\begin{equation}
 \hat{y}_{i,k} = \frac{\exp ( \tau {\mathbf f}_i^\top {\mathbf t}_k)}{\sum_j \exp ( \tau {\mathbf f}_i^\top {\mathbf t}_j)}
\end{equation} 

\paragraph{\textbf{Laplacian regularization}.} Laplacian regularizers are widely used in the context of graph/spectral clustering. This term encourages related samples (\textit{i.e.,} pairs of patches with high affinity $w_{i,j}$) to have similar label assignments. We build affinities based on the cosine similarities of each patch representation:

\begin{equation}
    w_{ij}= \mathbf{f}_i^\top\mathbf{f}_j
\end{equation}
In fact, affinity relations can be modified for each specific use-case, allowing to inject knowledge in the optimization process \cite{zanella2024test}. In our case, we can leverage the strong embedding capabilities of the image encoder to regularize the transductive procedure. In practice, to reduce memory needs, we sparsify the matrix by retaining only the 3 nearest neighbors of each patch.
\paragraph{\textbf{Objective function}.} 
We minimize the following objective:
\begin{equation}
\label{zero-shot-objective}
{\cal L}({\mathbf z}, \boldsymbol{\mu}, \boldsymbol{\Sigma}) =  \underbrace{- \frac{1}{|{\cal Q}|} \sum_{i\in \cal Q}  {\mathbf z}_{i}^\top \log ({\mathbf p}_{i})}_{\text{GMM clustering}} 
 \underbrace{-\sum_{i\in \cal Q} \sum_{j \in \cal Q} w_{ij} \mathbf{z}_{i}^\top\mathbf{z}_{j}}_{\text{Laplacian regularization}} + \underbrace{\sum_{i \in \cal Q} \mbox{KL} (\mathbf{z}_{i} || \hat{\mathbf{y}}_{i})}_{\text{Prediction penalty}}
\end{equation}
The Kullback–Leibler (KL) term encourages the prediction $\mathbf{z}_{i}$ not to deviate significantly from the zero-shot prediction $\hat{{\mathbf y}}_i$, thereby providing text supervision without the need of any labels.

\begin{algorithm}
  \caption{Histo-TransCLIP procedure for transductive inference alternates between \colorbox{green}{assignments} and \colorbox{red}{GMM-parameters} updates.\\ Input: $\mathbf{f}$ are the image embeddings, $\mathbf{t}$ are the text/class embeddings, $\tau$ is the temperature scaling used during each VLM pretraining.}
  \label{alg:transclip}
  \Function{Histo-TransCLIP($\mathbf{f}$, $\mathbf{t}$, $\tau$)}{
  \tcp{Text-based pseudo-labels $\hat{{\mathbf y}}$}
    $\hat{{\mathbf y}_i} =$ softmax$(\tau \mathbf{f}_i^T \mathbf{t})$ $\quad$ $\forall i$ \;
    \tcp{Initialize ${\mathbf z}$, ${\mathbf \mu}$, ${\mathbf \Sigma}$}
    ${\mathbf z}_i = \hat{{\mathbf y}_i}$ $\quad$ $\forall i$\;
    $\boldsymbol{\mu}_k =$ top\_confident\_average(${\mathbf f}$, $\hat{{\mathbf y}})$ $\quad$ $\forall k$\;
    $\text{diag}({\mathbf \Sigma}) = \frac{1}{n\_features}$\;
    \tcp{Iterative procedure}
    \While  {not\_converged} {  
        \boxitt{green}{2.}
        \For {l = 1:...}{
        $\mathbf{z}_{i}^{(l+1)} = \frac{\hat{\mathbf{y}}_{i} \odot \exp (\log (\mathbf{p}_{i}) + \sum_{j \in {\cal Q}} w_{ij} \mathbf{z}_{j}^{(l)})}{(\hat{\mathbf{y}}_{i} \odot \exp (\log (\mathbf{p}_{i}) + \sum_{j\in {\cal Q}} w_{ij}\mathbf{z}_{j}^{(l)}))^\top\mathbbm{1}_K}$ $\quad$ $\forall i$
        }
        \boxitt{red}{1.5}
        $\boldsymbol{\mu}_k = \frac{\sum_{i \in {\cal Q}} z_{i,k}  {\mathbf f}_i}{\sum_{i \in \cal Q} z_{i,k}}$$\quad$ $\forall k$\;
        $\text{diag}({\mathbf \Sigma}) = \frac{1}{|\cal Q|} \sum_{i \in \cal Q}\sum_{k} z_{i,k} (\mathbf{f}_i - \boldsymbol{\mu}_k)^2$\;
    }
    \Return{$\mathbf{z}$}\;
  }
\end{algorithm}

\paragraph{\textbf{Procedure.}} 
We refer to \cite{zanella2024boosting} for the technical details about the derivation. 
Optimizing \eqref{zero-shot-objective}, subject to simplex constraints,  
we obtain the following decoupled update rules for the assignment variables, which can be computed in parallel for all samples (\textit{i.e.,} patches) at a given iteration $l$:
\begin{align}
    \label{z-updates}
    \mathbf{z}_{i}^{(l+1)} = \frac{\hat{\mathbf{y}}_{i} \odot \exp (\log (\mathbf{p}_{i}) + \sum_{j \in {\cal D}} w_{ij} \mathbf{z}_{j}^{(l)})}{(\hat{\mathbf{y}}_{i} \odot \exp (\log (\mathbf{p}_{i}) + \sum_{j\in {\cal D}} w_{ij}\mathbf{z}_{j}^{(l)}))^\top\mathbbm{1}_K}
\end{align}
Note how each assignment $\mathbf{z}_{i}$ depends on its neighbors. This update must be computed iteratively until convergence, enabling assignments to propagate from the GMM likelihood to neighboring samples, weighted by their affinity. Since these updates are decoupled, this step can be parallelized efficiently (see runtime in Table \ref{tab:runtime}). With other variables fixed, we then have the following closed-form updates for the GMM parameters:   
\begin{align}
\label{u-update}
    \boldsymbol{\mu}_k = \frac{\sum_{i \in {\cal Q}} z_{i,k}  {\mathbf f}_i}{\sum_{i \in \cal Q} z_{i,k}}
\end{align}
\begin{align}
\label{sigma-update}
    \text{diag}({\mathbf \Sigma}) = \frac{1}{|\cal Q|} \sum_{i \in \cal Q}\sum_{k} z_{i,k} (\mathbf{f}_i - \boldsymbol{\mu}_k)^2
\end{align}
The procedure is summarized in Pseudocode \ref{alg:transclip} and alternates between solving~\eqref{z-updates} to get assignments  for each patch and computing the GMM parameters~\eqref{u-update}~\eqref{sigma-update} according to those assignments until convergence (see proof in \cite{zanella2024boosting}).

\begin{table*}[ht] 
\footnotesize
\centering
\renewcommand{\arraystretch}{1.05}
    \caption{Zero-shot and Histo-TransCLIP performance on top of various VLMs. Best values are highlighted in \textbf{bold}. \textcolor{DarkGreen}{\textbf{$\Delta_{\text{transductive}}$}} is the average accuracy gain brought by our transductive approach.}
    \label{table:mainresult}
    \vspace{0.1cm}
\begin{tabular}{llccccc}
\toprule
 Dataset & Method & \multicolumn{5}{c}{Model}\\
\midrule
\midrule
&  & CLIP& Quilt-B16 & Quilt-B32 & PLIP & CONCH \\
\midrule
\multirow{2}{*}{\textit{SICAP-MIL}} & Zero-shot & \textbf{29.85} & 40.44 & \textbf{35.04} & 46.84 & 27.71  \\
 & \cellcolor{LightBlue} Histo-TransCLIP & \cellcolor{LightBlue} 24.72 & \cellcolor{LightBlue} \textbf{58.49} & \cellcolor{LightBlue} 28.18 & \cellcolor{LightBlue} \textbf{53.23} & \cellcolor{LightBlue} \textbf{32.58}  \\
\midrule
\multirow{2}{*}{\textit{LC(Lung)}} & Zero-shot & \textbf{31.46} & 43.00 & 76.24 &  84.96 & 84.81  \\
 & \cellcolor{LightBlue} Histo-TransCLIP & \cellcolor{LightBlue} 25.62 & \cellcolor{LightBlue} \textbf{50.53} & \cellcolor{LightBlue} \textbf{93.93} & \cellcolor{LightBlue} \textbf{93.80} & \cellcolor{LightBlue} \textbf{96.29} \\
\midrule
\multirow{2}{*}{\textit{SKINCANCER}} & Zero-shot & 4.20 & 15.38 & 39.71 & 22.90 & 58.53  \\
& \cellcolor{LightBlue} Histo-TransCLIP & \cellcolor{LightBlue} \textbf{11.46} & \cellcolor{LightBlue} \textbf{33.33} & \cellcolor{LightBlue} \textbf{48.80} & \cellcolor{LightBlue} \textbf{36.72} & \cellcolor{LightBlue} \textbf{66.22}  \\
\midrule
\multirow{2}{*}{\textit{NCT-CRC}} & Zero-shot & 25.39 & 29.61 &  53.73 & 63.17 & 66.27  \\
& \cellcolor{LightBlue} Histo-TransCLIP & \cellcolor{LightBlue} \textbf{39.61} & \cellcolor{LightBlue} \textbf{48.40} & \cellcolor{LightBlue} \textbf{58.13} &  \cellcolor{LightBlue} \textbf{77.53} & \cellcolor{LightBlue} \textbf{70.36}  \\
\midrule
\multirow{3}{*}{\textit{Average}} & Zero-shot &  22.73 & 32.1 &  51.18 & 54.47 & 59.33 \\
& \cellcolor{LightBlue} Histo-TransCLIP &  \cellcolor{LightBlue} \textbf{25.35} & \cellcolor{LightBlue} \textbf{47.69} & \cellcolor{LightBlue} \textbf{57.26} & \cellcolor{LightBlue} \textbf{65.32} & \cellcolor{LightBlue} \textbf{66.36} \\
&  \textcolor{DarkGreen}{\textbf{$\Delta_{\text{transductive}}$}} & \textcolor{DarkGreen}{\textbf{+2.62}} & \textcolor{DarkGreen}{\textbf{+15.59}} & \textcolor{DarkGreen}{\textbf{+6.08}} & \textcolor{DarkGreen}{\textbf{+10.85}} & \textcolor{DarkGreen}{\textbf{+7.03}}\\
\bottomrule
\end{tabular}
\end{table*}

\section{Experiments}
\label{sec:experiments}

We conduct a comprehensive comparison of several vision-language models pretrained on histology images, namely PLIP \cite{plip}, QUILT \cite{ikezogwo2023quilt} (for which we report two versions) and CONCH \cite{lu2024avisionlanguage}. Text embeddings for each category are obtained following the specific 22 prompts used for CONCH (only one name is assigned to each target class), which are then averaged to get a single textual embedding per class. Numerical results are top-1 accuracy which compare zero-shot prediction (\textit{i.e.,} inductive inference) and Histo-TransCLIP (\textit{i.e.,} transductive inference).

\paragraph{\textbf{Datasets.}} We study different histopathology classification tasks on various organs and cancer types \cite{nct, sicap-mil, skincancer, lclung}. Specifically, \textit{NCT-CRC} \cite{nct} comprises patches of colorectal adenocarcinoma categorized into 9 classes, \textit{SICAP-MIL} \cite{sicap-mil} includes 4 prostate cancer grading, \textit{SKINCANCER} \cite{skincancer} is annotated with 9 anatomical tissue structures, and \textit{LC25000(Lung)}\cite{lclung} focuses on 3 classes of lung cancer. These diverse benchmarks enable the study of the generalization capability of VLMs pretrained on histology images and provide a thorough assessment of our transductive approach.

\paragraph{\textbf{Results.}} Table \ref{table:mainresult} presents a comparative analysis of zero-shot performance and the improvement achieved by Histo-TransCLIP. The lower classification accuracy of CLIP emphasizes the need for VLMs specifically pretrained on histology. Notably, the recently proposed CONCH model demonstrates the highest average accuracy. Note that the variation in zero-shot accuracies compared to the original paper values is largely influenced by the choice of prompt templates, for instance PLIP zero-shot results are significantly improved. This yields interesting questions on prompt sensitivy as discussed for future work in Section \ref{sec:conclusion}. Histo-TransCLIP consistently enhances performance significantly, highlighting the benefits of its transductive approach. Only in a few cases does the accuracy of Histo-TransCLIP drop, particularly when zero-shot performance is low due to direct regularization with initial text predictions. In most cases, Histo-TransCLIP effectively enhances performance, even on tasks initially achieving high accuracy, showcasing its strong ability to refine slightly misaligned text predictions for various VLMs.



\begin{table}[t]
    \centering
        \caption{Features denotes the runtime to pre-compute the image and text embeddings, Histo-TransCLIP denotes the runtime of our transductive procedure once embeddings are provided. Experiments were conducted on a single NVIDIA GeForce RTX 3090 (24Gb) GPU.}
    \label{tab:runtime}
    \vspace{0.1cm}
    \begin{tabular}{ccc}
        \toprule
         \#Patches & Features & Histo-TransCLIP \\
        \midrule
          $10^2$ & $\sim$ 1 sec. & $\sim$ 0.1 sec.\\
          $10^3$ & $\sim$ 4 sec. & $\sim$ 0.2 sec.\\
          $10^4$ & $\sim$ 28 sec. & $\sim$ 0.4 sec.\\
          $10^5$ & $\sim$ 5 min. & $\sim$ 6 sec.\\
         \bottomrule
       
    \end{tabular}
\end{table}
\paragraph{\textbf{Computational workload.}} Table \ref{tab:runtime} details the computational overhead associated with Quilt-B16 visual and textual feature extraction, alongside the implementation of Histo-TransCLIP across varying patch numbers in the \textit{NCT-CRC} dataset. While the time for feature extraction increases with the number of patches, the additional workload introduced by Histo-TransCLIP remains negligible. This shows transduction can importantly improve performance while maintaining black-box adaptation (\textit{i.e.,} without accessing the model's parameters) and without adding any notable additional workload.


\section{Conclusion}
\label{sec:conclusion}
We have demonstrated the significant value that transduction can bring to histology. By leveraging text-based predictions through a Kullback–Leibler divergence penalty and incorporating shared information among patches with Laplacian regularization, our approach significantly enhances the performance of vision-language models. Notably, our method is highly efficient and does not require additional labels or access to model parameters. 

\paragraph{\textbf{Future Work.}} Our approach can be naturally extended to the few-shot setting. Additionally, the quality of the prompts, \textit{i.e.,} the textual descriptions of each class, can significantly impact the final zero-shot performance. Studying this impact is undoubtedly valuable for safer applications. Finally, while our current work focuses on transduction using patches from multiple slides, a more constrained and valuable application would involve transduction on patches from a single slide to improve performance on a per-patient basis.

\begin{credits}
\subsubsection{\ackname}
M.~Zanella is funded by the Walloon region under grant No.~2010235 (ARIAC by DIGITALWALLONIA4.AI). F.~Shakeri is funded by Natural Sciences and Engineering Research Council of Canada (NSERC) and Canadian Institutes of Health Research (CIHR).

\subsubsection{\discintname} The authors have no competing interests to declare that are relevant to the content of this article

\end{credits}

\bibliographystyle{splncs04}
\bibliography{biblio}


\end{document}